\documentclass[conference]{IEEEtran}
\IEEEoverridecommandlockouts

\usepackage{cite}
\usepackage{amsmath,amssymb,amsfonts}
\usepackage{algorithmic}
\usepackage{graphicx}
\usepackage{textcomp}
\usepackage{orcidlink}
\usepackage{xcolor}
\usepackage{subcaption}
 \usepackage{multirow}
\def\BibTeX{{\rm B\kern-.05em{\sc i\kern-.025em b}\kern-.08em
    T\kern-.1667em\lower.7ex\hbox{E}\kern-.125emX}}
\begin{document}

\title{BLENDS: Bayesian Learning-Enhanced Deep Smoothing for GNSS-Denied Environments}

\author{\IEEEauthorblockN{1\textsuperscript{st} Nadav Cohen~\orcidlink{0000-0002-8249-0239}}
\IEEEauthorblockA{\textit{The Hatter Department of Marine Technologies} \\
\textit{University of Haifa}\\
Haifa, Israel \\
ncohe140@campus.haifa.ac.il}
\and
\IEEEauthorblockN{2\textsuperscript{nd} Itzik Klein~ \orcidlink{0000-0001-7846-0654} }
\IEEEauthorblockA{\textit{The Hatter Department of Marine Technologies} \\
\textit{University of Haifa}\\
Haifa, Israel \\
kitzik@univ.haifa.ac.il}

}

\maketitle

\begin{abstract}
Maintaining accurate navigation during GNSS outages remains a significant challenge for autonomous systems relying on low-cost inertial sensors. While classical smoothing methods, such as the two-filter smoother and Rauch–Tung–Striebel smoother, exploit measurements collected before and after an outage, their performance remains limited by the accuracy of conventional GNSS measurements. This paper presents Bayesian learning-enhanced navigation with deep smoothing (BLENDS), a transformer-based framework that augments Bayesian smoothing with learned covariance adaptation and state correction. The proposed method preserves the statistical foundations of Bayesian estimation while leveraging data-driven learning to improve navigation accuracy. Evaluations on the quadrotor dataset with GNSS outages demonstrate that BLENDS consistently outperforms both model-based smoothers, achieving up to 25.6\% improvement in the position root mean square error while also reducing estimation uncertainty. Furthermore, BLENDS learns to compensate for the systematic bias between conventional GNSS positioning and RTK ground truth, enabling accuracy beyond that achievable with conventional GNSS measurements alone. The results demonstrate the potential of learning-enhanced Bayesian smoothing for resilient and high-accuracy navigation in GNSS-challenged environments.
\end{abstract}


\section{Introduction}
\noindent   
\IEEEPARstart{N}{avigation} precision is a critical aspect of deploying autonomous robotic systems. With the growing interest in using platforms such as drones, small ground vehicles, marine surface vehicles, and others, achieving accurate navigation using low-cost, affordable sensors remains a significant challenge. The most basic and commonly used approach for obtaining a navigation solution is the integration of an inertial navigation system (INS) with a global navigation satellite system (GNSS), as their combination mitigates the individual limitations of each system. The INS, by itself, can provide a complete navigation solution consisting of the platform's position, velocity, and orientation in a referenced coordinate system. However, due to its dead-reckoning nature and inherent sensor errors, it tends to accumulate errors over time. GNSS, on the other hand, provides reliable and accurate position measurements that mitigate inertial drift, but it does not directly provide orientation measurements~\cite{farrell2008aided}.
\\ \noindent
The most common approach for integrating INS and GNSS is through the extended Kalman filter (EKF), which addresses the nonlinear nature of the problem and minimizes estimation uncertainty~\cite{titterton2004strapdown,borko2018gnss,silva2025dual}. However, when GNSS signals are blocked or unavailable, the EKF effectively acts as a mathematical integrator, causing errors to accumulate over time. In addition, GNSS alone typically provides position estimates with errors ranging from 2 to 5 meters, which may be insufficient for many modern applications requiring high precision. Therefore, real-time kinematic (RTK) corrections can be employed to improve positioning performance to centimeter-level accuracy. However, such solutions are generally expensive and contradict the demand for low-cost and affordable navigation systems.
\\ \noindent
Solutions for GNSS outages can be found in the implementation of post-processing algorithms such as the Rauch--Tung--Striebel smoother (RTSS)~\cite{rauch1965maximum} and the two-filter smoother (TFS)~\cite{fraser1969optimum}. These methods utilize information from the entire dataset, including measurements acquired both before and after the outage, to smooth and refine the navigation solution. Although they have demonstrated promising results in such applications, their performance remains limited by the accuracy of conventional GNSS positioning, which typically exhibits errors on the order of a few meters. Considerable attention has been given to developing techniques that mitigate the impact of GNSS outages and signal degradation on navigation accuracy. Classical smoothing techniques have been shown to reduce estimation errors and improve solution robustness in GNSS-challenged environments~\cite{liu2010two,zhang2017new,yin2023sensor,zhao2025smoothing}. At the same time, the growing availability of computational resources has accelerated the adoption of machine learning methods in navigation systems. Recent studies have explored learning-based enhancements to model-based estimators, including the prediction of filter parameters, adaptation of uncertainty models, and reconstruction of missing measurements~\cite{revach2022kalmannet,wu2020predicting,cohen2025adaptive,levy2026adaptive,fang2020lstm,cohen2024seamless}. Learning-based smoothing has also been investigated through architectures such as RTSNet~\cite{ni2022rtsnet,revach2023rtsnet}. Nevertheless, these approaches have not been specifically developed for high-precision INS/GNSS smoothing and do not directly address the positioning errors inherent in conventional GNSS measurements.
\\ \noindent
In our recent work, we introduced BLENDS, a Bayesian learning-enhanced smoothing framework that integrates deep learning within the classical smoothing process to correct the systematic position bias inherited from conventional GNSS measurements while preserving the theoretical foundations of Bayesian estimation~\cite{cohen2026bayesian}. The results demonstrated that learning-based smoothing can improve positioning accuracy beyond the limits imposed by standard GNSS measurements. Motivated by these findings, this paper extends the concept to GNSS outage scenarios, where the objective is not only to bridge GNSS gaps and mitigate navigation drift, but also to enhance the overall navigation solution beyond the accuracy achievable with conventional GNSS positioning. Experimental results demonstrate that the proposed approach consistently outperforms classical smoothing methods, achieving substantial reductions in both navigation error and estimation uncertainty during GNSS outages while maintaining the statistical consistency of the underlying Bayesian framework.
\\ \noindent
The remainder of this paper is organized as follows. Section~\ref{sec:problem} presents the problem formulation and reviews the EKF, TFS, and RTSS. Section~\ref{sec:proposed} introduces BLENDS. Section~\ref{sec:results} describes the dataset, evaluation metrics, implementation details, and GNSS outage simulation methodology, and presents the experimental results. Finally, Section~\ref{sec:conclusion} concludes the paper and outlines directions for future research.

\section{Problem Formulation}\label{sec:problem}
\begin{figure*}[t]
    \centering
    \includegraphics[width=0.8\textwidth]{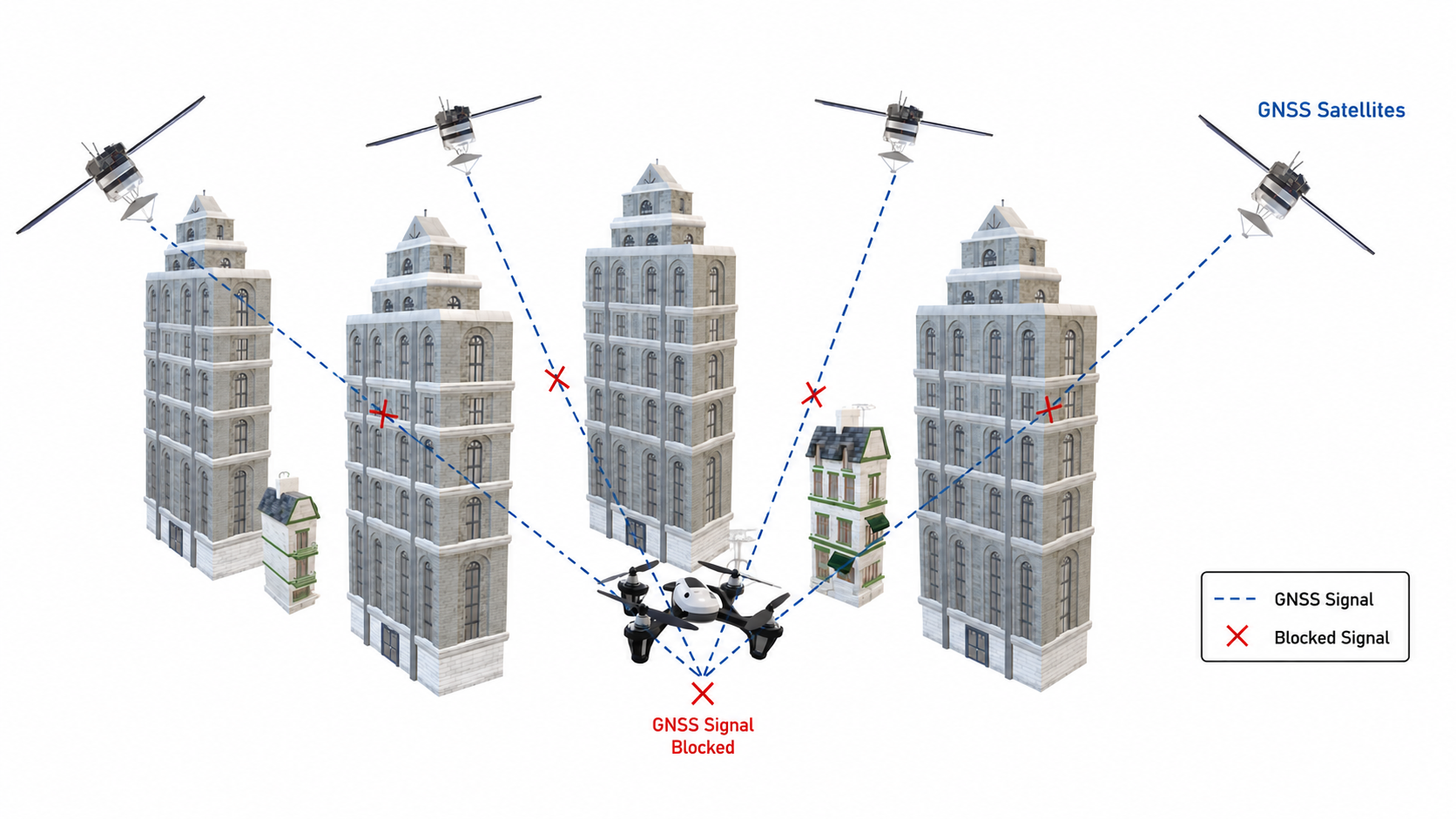}
    \caption{Urban GNSS-denied scenario considered in this work. High-rise buildings block the direct propagation of GNSS signals from multiple satellites to the quadrotor, creating signal outages.}
    \label{fig:urban_gnss_blockage}
\end{figure*}

\noindent
During nominal operation, the navigation solution is obtained through a loosely coupled INS/GNSS integration framework based on an error-state EKF. The filter estimates the forward error-state vector
\begin{equation}
\delta\boldsymbol{x}^{n}_{f} = \left[ (\delta\boldsymbol{p}^{n})^{T},\ (\delta\boldsymbol{v}^{n})^{T},\ (\delta\boldsymbol{\epsilon}^{n})^{T},\ (\delta\boldsymbol{b}_{a})^{T},\ (\delta\boldsymbol{b}_{g})^{T} \right]^{T},
\end{equation}
where $\delta\boldsymbol{p}^{n}$, $\delta\boldsymbol{v}^{n}$, $\delta\boldsymbol{\epsilon}^{n}$, $\delta\boldsymbol{b}_{a}$, and $\delta\boldsymbol{b}_{g}$ denote the position, velocity, orientation, accelerometer bias, and gyroscope bias errors, respectively. The superscript $n$ denotes the navigation frame and the subscript $f$ denotes the forward filter. The error-state vector and its associated covariance matrix $\mathbf{P}_{f}$ are propagated according to
\begin{equation}
\delta\boldsymbol{x}^{-}_{f,k} = \mathbf{\Phi}_{k-1}\, \delta\boldsymbol{x}^{+}_{f,k-1},
\end{equation}
\begin{equation}
\mathbf{P}^{-}_{f,k} = \mathbf{\Phi}_{k-1}\, \mathbf{P}^{+}_{f,k-1}\, \mathbf{\Phi}_{k-1}^{T} + \mathbf{Q}_{k-1},
\end{equation}
where $\mathbf{\Phi}_{k-1}$ is the state-transition matrix and $\mathbf{Q}_{k-1}$ is the process noise covariance matrix. Whenever a GNSS measurement is available, the filter computes the measurement residual
\begin{equation}
\delta\boldsymbol{z}_{k} = \boldsymbol{p}^{\mathrm{INS}}_{k} - \boldsymbol{p}^{\mathrm{GNSS}}_{k},
\end{equation}
where $\boldsymbol{p}^{\mathrm{INS}}_{k}$ and $\boldsymbol{p}^{\mathrm{GNSS}}_{k}$ are the INS-predicted and GNSS-measured positions, respectively. The correction step is then performed as
\begin{equation}
\delta\boldsymbol{x}^{+}_{f,k} = \delta\boldsymbol{x}^{-}_{f,k} + \mathbf{K}_{k} \left( \delta\boldsymbol{z}_{k} - \mathbf{H}_{k}\, \delta\boldsymbol{x}^{-}_{f,k} \right),
\end{equation}
where $\mathbf{K}_{k}$ is the Kalman gain and $\mathbf{H}_{k}$ is the measurement matrix. The complete error-state EKF formulation for INS/GNSS can be found in~\cite{groves2015principles}.
\\ \noindent
However, when the platform enters a GNSS-denied environment, GNSS measurements become unavailable and the correction step can no longer be performed. Consequently, the EKF relies solely on the prediction stage, causing navigation errors to accumulate over time due to inertial sensor noise and bias drift. An illustration of a GNSS-denied environment in an urban setting is shown in Fig.~\ref{fig:urban_gnss_blockage}.
\\ \noindent
Unlike real-time navigation, post-processing approaches have access to measurements
collected after the GNSS outage has occurred. Consequently, information from both
past and future observations can be exploited to improve the navigation solution. A
common approach for achieving this objective is the TFS, which combines the estimates of a forward error-state EKF and a backward information filter.
\\ \noindent
The forward filter operates in run, while the backward
filter processes the same dataset in reverse chronological order, beginning from the
final epoch. To ensure independence between the forward and backward estimates, the
backward filter is initialized with no prior information according to
\begin{equation}
\mathbf{P}_{b,N}^{-} \to \infty, \qquad
\mathbf{\mathcal{I}}_{b,N}^{-} = \left(\mathbf{P}_{b,N}^{-}\right)^{-1} = \mathbf{0},
\end{equation}
where $\mathbf{P}_{b,N}^{-}$ and $\mathbf{\mathcal{I}}_{b,N}^{-}$ denote the backward
covariance and information matrices at the final epoch $N$, respectively. The backward
filter then propagates information backward through time, producing the backward
error-state estimate $\delta\boldsymbol{x}_{b,k}$ and its associated covariance matrix
$\mathbf{P}_{b,k}$.
\\ \noindent
The TFS combines the forward and backward information according to
\begin{equation}
\delta\boldsymbol{x}_{s,k} =
\mathbf{P}_{s,k} \left(
    \mathbf{\mathcal{I}}_{f,k}\, \delta\boldsymbol{x}_{f,k}
    + \mathbf{\mathcal{I}}_{b,k}\, \delta\boldsymbol{x}_{b,k}
\right),
\end{equation}
with the corresponding smoothed covariance given by
\begin{equation}
\mathbf{P}_{s,k} = \left(
    \mathbf{\mathcal{I}}_{f,k} + \mathbf{\mathcal{I}}_{b,k}
\right)^{-1},
\end{equation}
where $\delta\boldsymbol{x}_{s,k}$ and $\mathbf{P}_{s,k}$ denote the smoothed
error-state estimate and covariance matrix, respectively, and
$\mathbf{\mathcal{I}}_{f,k} = \mathbf{P}_{f,k}^{-1}$ and
$\mathbf{\mathcal{I}}_{b,k} = \mathbf{P}_{b,k}^{-1}$ are the forward and backward
information matrices. The TFS therefore provides an optimal minimum-variance estimate
by fusing information from measurements collected both before and after the GNSS
outage. However, although smoothing can significantly reduce estimation uncertainty
and navigation drift, its performance remains fundamentally limited by the accuracy
of the available GNSS measurements.
\\ \noindent
Another widely used fixed-interval smoothing approach is the RTSS. Unlike the TFS, which combines independent forward and backward
estimates, the RTSS operates directly on the stored outputs of the forward EKF.
Specifically, it utilizes the predicted and updated error-state estimates,
$\delta\boldsymbol{x}^{-}_{f,k}$ and $\delta\boldsymbol{x}^{+}_{f,k}$, together
with their corresponding covariance matrices, $\mathbf{P}^{-}_{f,k}$ and
$\mathbf{P}^{+}_{f,k}$, to recursively smooth the navigation solution backward in
time. The RTSS smoothing equations are given by
\begin{equation}
\mathbf{K}_{s,k} = \mathbf{P}^{+}_{f,k}\,\mathbf{\Phi}_{k}^{T}
\left(\mathbf{P}^{-}_{f,k+1}\right)^{-1},
\end{equation}
\begin{equation}
\mathbf{P}_{s,k} = \mathbf{P}^{+}_{f,k} + \mathbf{K}_{s,k}
\left(\mathbf{P}_{s,k+1} - \mathbf{P}^{-}_{f,k+1}\right)\mathbf{K}_{s,k}^{T},
\end{equation}
\begin{equation}
\delta\boldsymbol{x}_{s,k} = \delta\boldsymbol{x}^{+}_{f,k} + \mathbf{K}_{s,k}
\left(\delta\boldsymbol{x}_{s,k+1} - \delta\boldsymbol{x}^{-}_{f,k+1}\right),
\end{equation}
where $\mathbf{K}_{s,k}$ denotes the smoother gain matrix, while
$\delta\boldsymbol{x}_{s,k}$ and $\mathbf{P}_{s,k}$ represent the smoothed
error-state estimate and covariance matrix, respectively. Similar to the TFS, the
RTSS exploits measurements collected both before and after a GNSS outage to refine
the navigation solution. Furthermore, under the linear Gaussian assumption, the TFS
and RTSS are theoretically equivalent and yield identical smoothed estimates, although
they differ in their computational implementation.

\section{Proposed Approach}\label{sec:proposed}
\begin{figure}[h!]
    \centering
    \includegraphics[width=0.6\columnwidth]{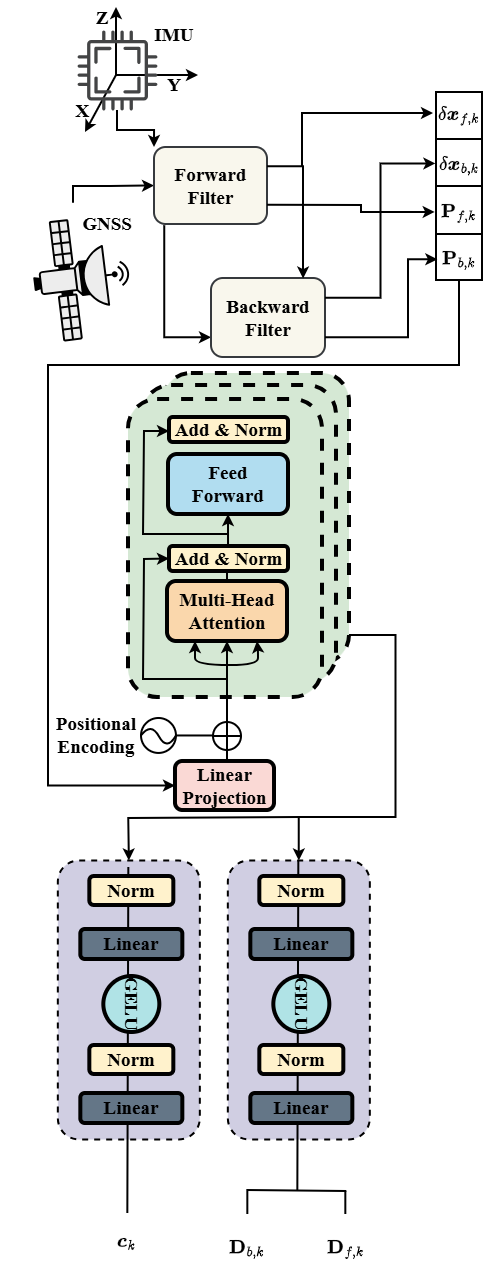}
    \caption{Overview of the proposed BLENDS framework for GNSS-denied environments. The outputs of the forward and backward smoothing filters are processed by a transformer network, which learns to refine the smoothing process and improve the final navigation solution.}

    \label{fig:archi}
\end{figure}

\noindent
To both mitigate navigation degradation during GNSS outages and improve overall accuracy, we propose a Bayesian learning-enhanced navigation with deep smoothing framework. It augments the classical TFS with a neural network that learns to adapt the smoothing process from data while preserving the underlying Bayesian estimation framework.
\\ \noindent
The network receives as input the forward and backward error-state estimates together
with their associated covariance matrices:
\begin{equation}
\boldsymbol{u}_{k} = \left[
    \delta\boldsymbol{x}^T_{f,k},\;
    \delta\boldsymbol{x}^T_{b,k},\;
    \mathrm{vec}(\mathbf{P}_{f,k})^T,\;
    \mathrm{vec}(\mathbf{P}_{b,k})^T
\right]^T.
\end{equation}
A transformer-based network processes the complete trajectory and produces three
outputs at each epoch:
\begin{equation}
\mathbf{D}_{f,k},\; \mathbf{D}_{b,k},\; \boldsymbol{c}_{k}
= \mathcal{F}_{\boldsymbol{\theta}}(\boldsymbol{u}_{1:N}),
\end{equation}
where $\mathbf{D}_{f,k}$ and $\mathbf{D}_{b,k}$ are covariance scaling matrices, and
$\boldsymbol{c}_{k}$ is an additive correction term. The scaling matrices modify the
forward and backward covariance matrices according to
\begin{equation}
\tilde{\mathbf{P}}_{f,k} = \mathbf{D}_{f,k}\,\mathbf{P}_{f,k}\,\mathbf{D}_{f,k}^{T},
\qquad
\tilde{\mathbf{P}}_{b,k} = \mathbf{D}_{b,k}\,\mathbf{P}_{b,k}\,\mathbf{D}_{b,k}^{T},
\end{equation}
yielding learned information weights within the TFS framework. The correction term
$\boldsymbol{c}_{k}$ additionally enables direct compensation of systematic errors
that cannot be addressed through covariance scaling alone.
\\ \noindent
The resulting BLENDS smoothed estimate is computed as
\begin{equation}
\delta\boldsymbol{x}^{\mathrm{BLENDS}}_{s,k} =
\tilde{\mathbf{P}}_{s,k}
\left(
    \tilde{\mathbf{\mathcal{I}}}_{f,k}\,\delta\boldsymbol{x}_{f,k}
    + \tilde{\mathbf{\mathcal{I}}}_{b,k}\,\delta\boldsymbol{x}_{b,k}
\right) + \boldsymbol{c}_{k},
\end{equation}
where $\tilde{\mathbf{P}}_{s,k} = \left(\tilde{\mathbf{\mathcal{I}}}_{f,k} +
\tilde{\mathbf{\mathcal{I}}}_{b,k}\right)^{-1}$ denotes the learned smoothed
covariance, and $\tilde{\mathbf{\mathcal{I}}}_{f,k} = \tilde{\mathbf{P}}_{f,k}^{-1}$
and $\tilde{\mathbf{\mathcal{I}}}_{b,k} = \tilde{\mathbf{P}}_{b,k}^{-1}$ are the
scaled information matrices, respectively.
\\ \noindent
To ensure consistency with Bayesian estimation theory, the network is trained using
a Bayesian-consistent loss (BCL), defined as
\begin{equation}
\mathcal{L}_{\mathrm{BCL}} = \frac{1}{BT}
\sum_{b=1}^{B} \sum_{k=1}^{T}
\left(
    \lambda_{p}\,\mathcal{L}^{(p)}_{k}
    + \lambda_{v}\,\mathcal{L}^{(v)}_{k}
    + \lambda_{r}\,\mathcal{L}^{(R)}_{k}
    + \lambda_{c}\,\mathcal{L}^{(c)}_{k}
\right)
\end{equation}
where the first three terms supervise the position, velocity, and orientation estimates,
respectively, while the final term penalizes the trace of the smoothed covariance
matrix,
\begin{equation}
\mathcal{L}^{(c)}_{k} = \mathrm{Tr}\!\left(\mathbf{P}_{s,k}\right),
\end{equation}
encouraging minimum-variance solutions. Consequently, BLENDS learns to exploit
information available both before and after the GNSS outage while maintaining
statistically consistent state and covariance estimates.
The overall BLENDS architecture is illustrated in Fig.~\ref{fig:archi}, where the outputs of the forward and backward filters are jointly processed by a transformer network to produce learned covariance scaling matrices and an additive correction term, which are subsequently incorporated into the smoothing framework to obtain the final navigation solution.
More in depth description of the BLENDS pipeline can be found in the original paper~\cite{cohen2026bayesian}.
\section{Analysis and Results}\label{sec:results}
\subsection{Dataset}\label{sec:dataset}
\noindent
The proposed approach is evaluated using the INSANE benchmark
dataset~\cite{brommer2022insane}, collected with an aerial quadrotor platform
equipped with a Pixhawk4 autopilot and an external RTK system providing
centimeter-level ground truth (GT) positioning. This work utilizes the Mars analogue
desert subset acquired at the Ramon Crater in Israel, which offers realistic flight
conditions and diverse motion patterns representative of autonomous aerial navigation.
\\ \noindent
The onboard inertial measurements are provided by an ICM-20689 MEMS IMU operating
at 200\,Hz, while GNSS measurements are available at 5\,Hz. GT position
estimates are obtained from the RTK system and recorded at 8\,Hz. The ICM-20689 is a low-cost consumer-grade MEMS
IMU characterized by higher noise levels and greater bias instability, making
long-term inertial navigation particularly challenging during GNSS outages.
\\ \noindent
Following data quality inspection, 13 trajectories were selected from the original dataset. The training set consists of approximately 23.7 minutes of flight data taken for two out of three flight zones. Two trajectories, denoted as Trajectory~1 and Trajectory~2, are reserved exclusively for testing and were not used during training. These trajectories comprise approximately 4.8 minutes of flight data in total. Notably, one of the test trajectories originates from the unseen Zone~3, providing a challenging evaluation of the proposed method's ability to generalize to new environments and motion patterns.
Table~\ref{tab:dataset_summary} summarizes the main characteristics of the dataset
and the corresponding train, and test splits.
\begin{table}[h!]
\centering
\caption{Summary of the INSANE quadrotor dataset used in this work.}
\label{tab:dataset_summary}
\begin{tabular}{lc}
\hline
 & Quadrotor \\
\hline
Platform & Pixhawk4 Quadrotor \\
IMU Rate [Hz] & 200 \\
GNSS Rate [Hz] & 5 \\
GT Source & RTK \\
GT Rate [Hz] & 8 \\
Training [min] & 23.7 \\
Testing [min] & 4.8 \\
Number of Trajectories & 13 \\
Environment & Desert \\
\hline
\end{tabular}
\end{table}

\noindent
To evaluate the proposed method under GNSS-denied conditions, an artificial GNSS
outage of 10 seconds was introduced into every test trajectory. During
each outage interval, GNSS measurements were withheld from the EKF, forcing the
navigation solution to rely solely on inertial propagation and consequently leading
to the accumulation of navigation errors. The outage locations were selected randomly.
\subsection{Performance Metrics}
\noindent
The performance of each estimator is evaluated using the position root mean square
error (PRMSE), defined as
\begin{equation}
\text{PRMSE} = \sqrt{ \frac{1}{T} \sum_{k=1}^{T} \left\| \boldsymbol{e}^{p}_{k} \right\|^{2} },
\label{eqn:prmse}
\end{equation}
where $\boldsymbol{e}^{p}_{k}$ denotes the position error vector at time step $k$
and $T$ is the total number of evaluated samples. The PRMSE is reported in meters
and quantifies the overall deviation of the estimated trajectory from the GT trajectory. Additionally, to quantify the reduction in estimation uncertainty, the percent
covariance improvement (PCI) is defined as
\begin{equation}
\text{PCI}_{k} = 100 \times
\frac{ \mathrm{Tr}\!\left( \mathbf{P}^{\mathrm{ref}}_{k} \right) -
       \mathrm{Tr}\!\left( \mathbf{P}^{\mathrm{test}}_{k} \right) }
     { \mathrm{Tr}\!\left( \mathbf{P}^{\mathrm{ref}}_{k} \right) },
\label{eqn:pci}
\end{equation}
where $\mathbf{P}^{\mathrm{ref}}_{k}$ and $\mathbf{P}^{\mathrm{test}}_{k}$ denote
the covariance matrices of the reference and evaluated estimators, respectively. In
this work, the forward EKF serves as the reference estimator. A positive PCI value
indicates a reduction in estimation uncertainty relative to the forward EKF, with
larger values reflecting greater effectiveness of the smoothing algorithm.
\subsection{Implementation Details}
\noindent
The proposed network is implemented in PyTorch and follows the same architecture, training procedure, and hyperparameter configuration introduced in the original BLENDS framework~\cite{cohen2026bayesian}. In particular, the output layers responsible for the covariance scaling matrices and additive correction term are initialized to zero, ensuring that the network reduces to the standard TFS at initialization. GT labels are generated by applying the TFS to the inertial and RTK measurements, yielding a smoothed reference trajectory for supervised learning. Unless otherwise stated, all training hyperparameters are identical to those reported in the original BLENDS paper.
\subsection{Results}
\noindent
The proposed BLENDS framework was evaluated on two previously unseen trajectories
from the INSANE dataset, denoted Trajectory~1 and Trajectory~2.
Table~\ref{tab:rmse} summarizes the PRMSE improvement relative to the EKF
baseline. While the classical TFS and RTSS provide only marginal improvements and
occasionally degrade performance, BLENDS consistently achieves the largest reduction
in position error across all axes and both trajectories. This outcome is expected because, although the model-based approaches are able to compensate for GNSS outages, they cannot correct the inherent bias between standard GNSS measurements and the RTK reference solution. In contrast, BLENDS is trained not only to bridge GNSS gaps but also to compensate for this bias, resulting in improved positioning accuracy relative to the RTK GT.
\\ \noindent
For Trajectory~1, BLENDS reduces the North and East PRMSE by 32.1\% and 33.3\%,
respectively, resulting in a 16.7\% improvement in the overall PRMSE. In
comparison, TFS and RTSS provide negligible improvements. A similar
trend is observed for Trajectory~2, where BLENDS achieves improvements of 66.4\%,
21.3\%, and 16.5\% in the North, East, and Down directions, respectively,
corresponding to a 25.6\% reduction in the overall PRMSE.

\begin{figure}[h!]
\centering
\begin{subfigure}{0.48\columnwidth}
    \includegraphics[width=\linewidth]{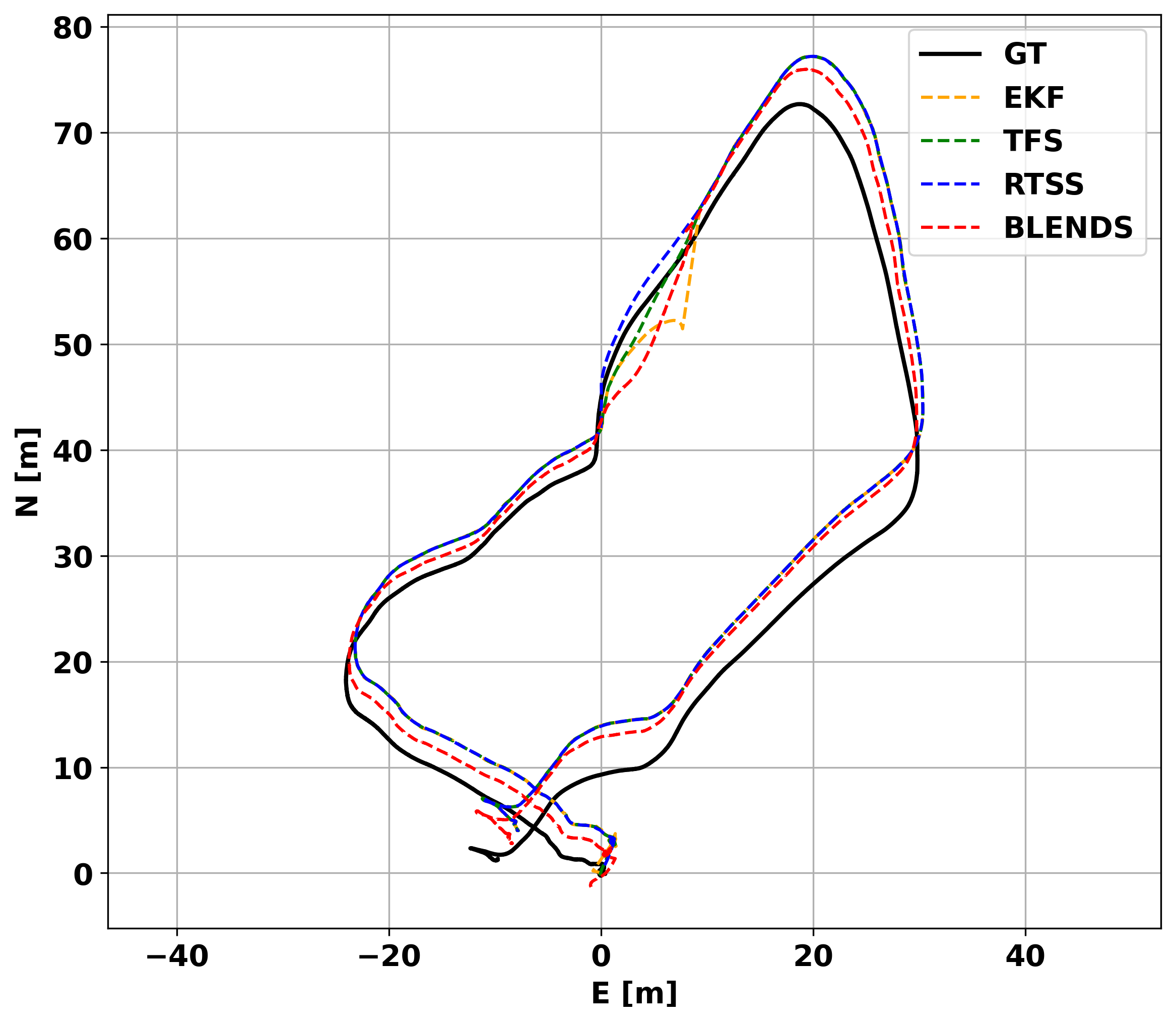}
    \caption{Trajectory 1.}
\end{subfigure}
\hfill
\begin{subfigure}{0.48\columnwidth}
    \includegraphics[width=\linewidth]{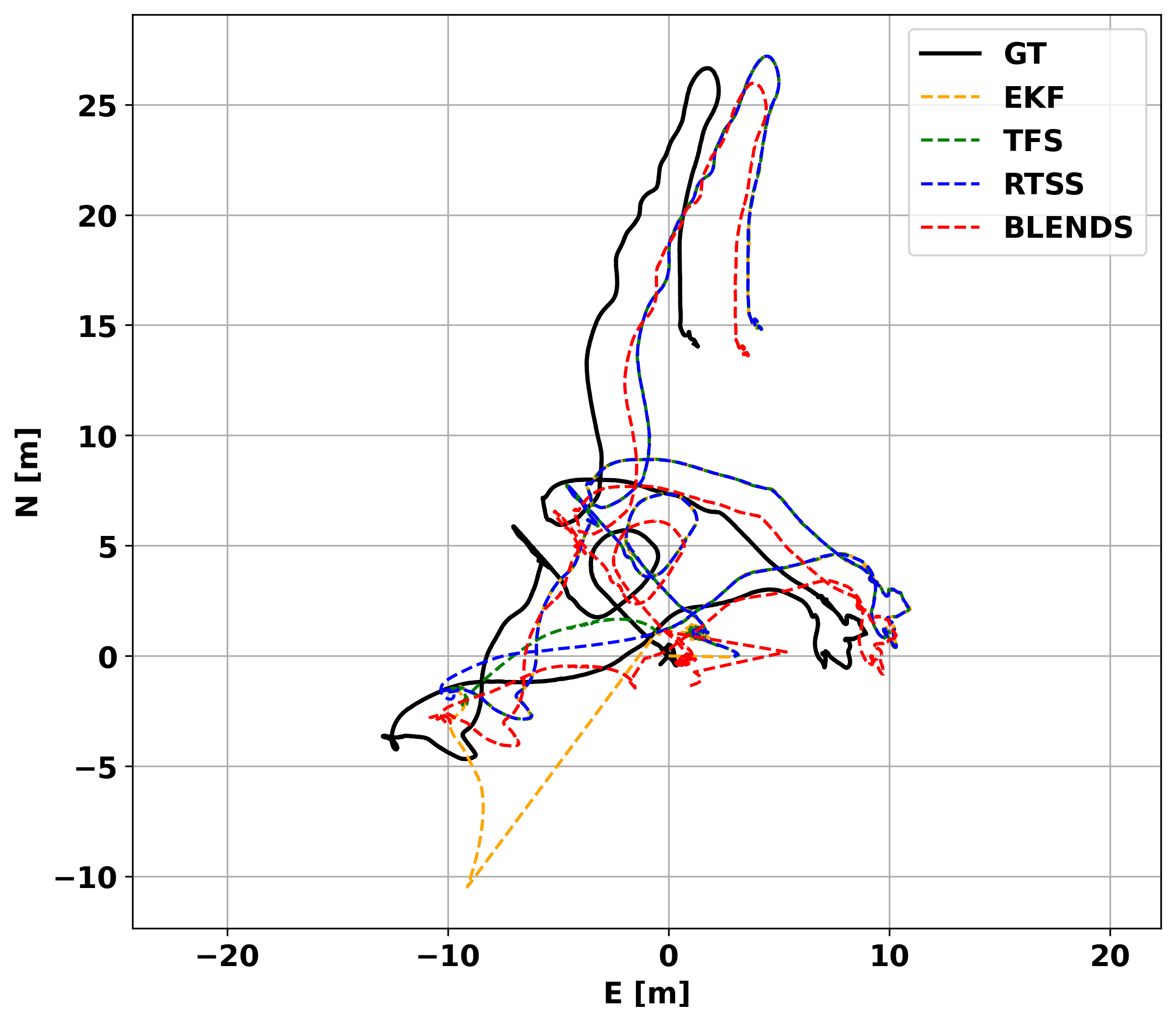}
    \caption{Trajectory 2.}
\end{subfigure}
\caption{Trajectory comparison between the ground truth and the estimated
solutions produced by the EKF, TFS, RTSS, and BLENDS.}
\label{fig:traj_results}
\end{figure}
\begin{figure}[h!]
\centering
\begin{subfigure}{0.48\columnwidth}
    \includegraphics[width=\linewidth]{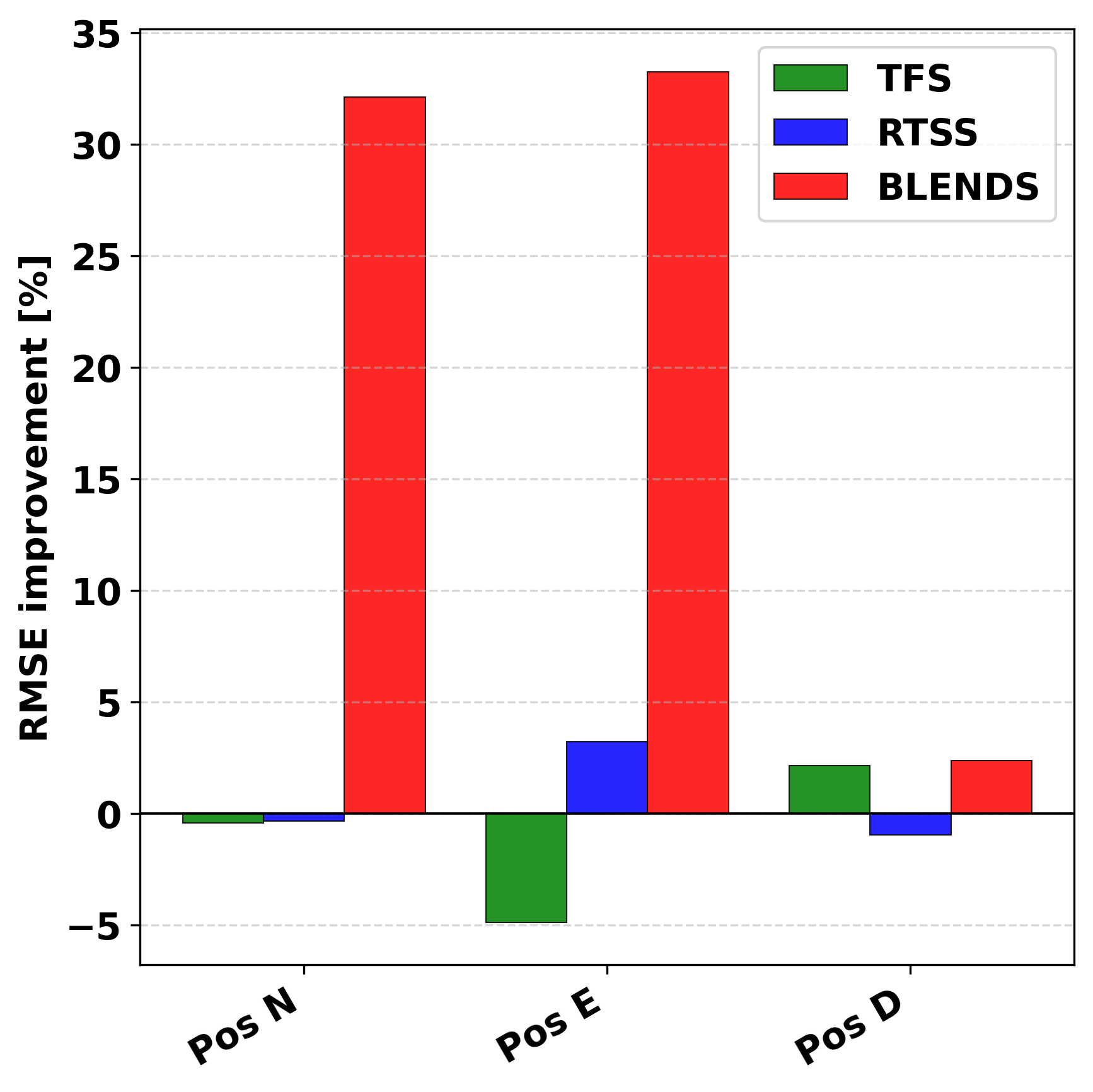}
    \caption{Trajectory 1.}
\end{subfigure}
\hfill
\begin{subfigure}{0.48\columnwidth}
    \includegraphics[width=\linewidth]{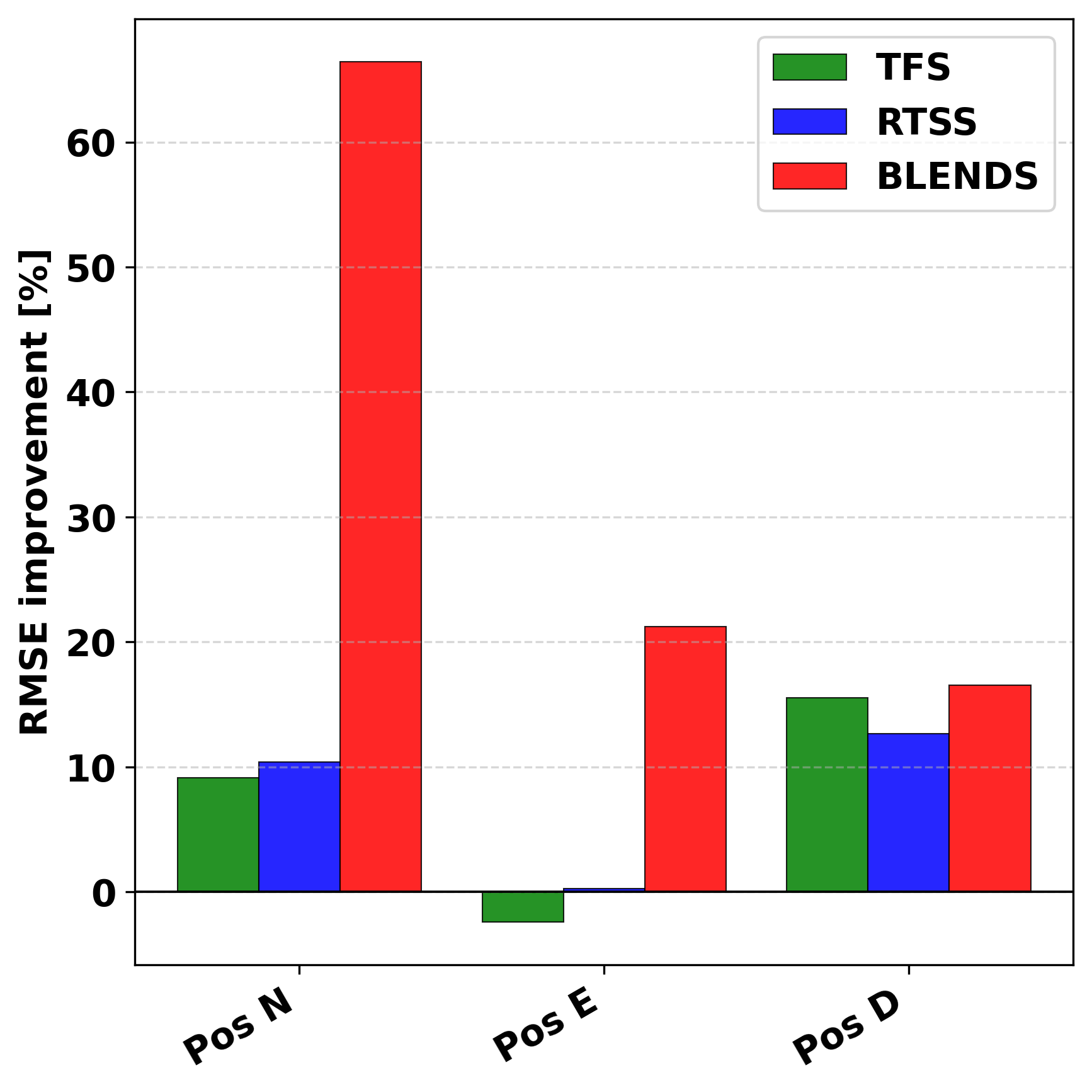}
    \caption{Trajectory 2.}
\end{subfigure}
\caption{PRMSE improvement relative to the EKF baseline for the North, East, and
Down position components. Positive values indicate improved performance compared
with the EKF.}
\label{fig:prmse_results}
\end{figure}
\begin{figure*}[h!]
\centering
\begin{subfigure}{0.48\textwidth}
    \includegraphics[width=\linewidth]{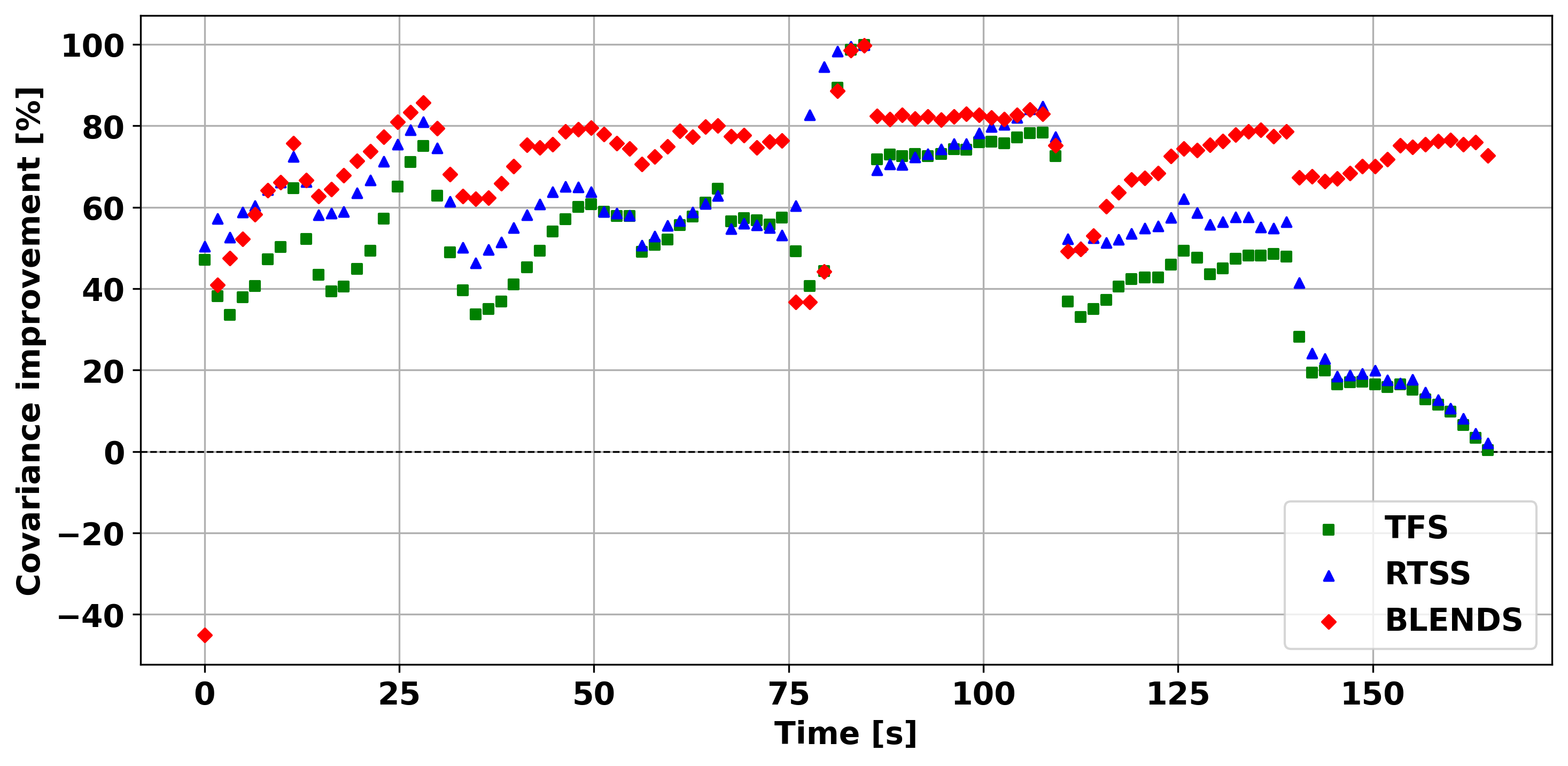}
    \caption{Trajectory 1.}
\end{subfigure}
\hfill
\begin{subfigure}{0.48\textwidth}
    \includegraphics[width=\linewidth]{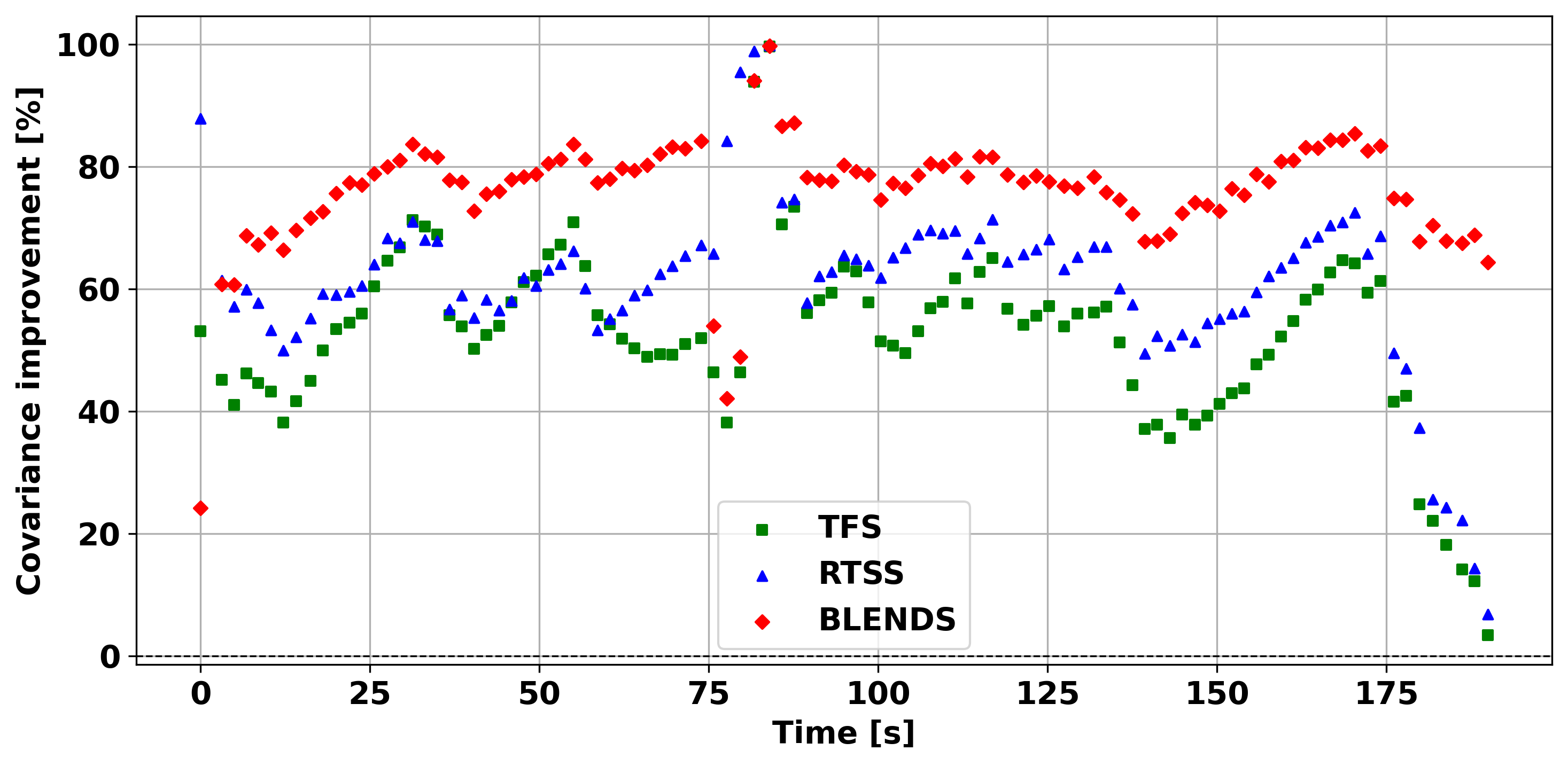}
    \caption{Trajectory 2.}
\end{subfigure}
\caption{Percent covariance improvement relative to the EKF covariance.
Higher values indicate a greater reduction in estimation uncertainty.}
\label{fig:pci_results}
\end{figure*}

\noindent
The trajectory comparisons shown in Fig.~\ref{fig:traj_results} support these
quantitative results. Although TFS and RTSS produce nearly identical trajectories,
BLENDS exhibits a noticeably closer agreement with the GT trajectory, particularly
in regions affected by the GNSS outage. This behavior is further confirmed by the
per-axis PRMSE improvements presented in Fig.~\ref{fig:prmse_results}.
\\ \noindent
In addition to improving positioning accuracy, BLENDS also achieves the largest
reduction in estimation uncertainty. As shown in Fig.~\ref{fig:pci_results}, the
proposed method consistently attains higher PCI values than both TFS and RTSS
throughout the trajectories, typically maintaining covariance reductions of
70--85\%. These results indicate that the learned covariance adaptation effectively
exploits future information while preserving the statistical consistency of the
smoothing framework.

\begin{table}[t]
\centering
\caption{PRMSE [m] for the evaluated methods.}
\label{tab:rmse}
\begin{tabular}{llcccc}
\hline
Trajectory & Method & N [m] & E [m] & D [m] & 3D [m] \\
\hline
\multirow{4}{*}{Trajectory 1}
 & EKF    & 3.695          & 1.134          & 3.837          & 5.289          \\
 & TFS    & 3.711          & 1.189          & 3.755          & 5.293          \\
 & RTSS   & 3.708          & 1.097          & 3.874          & 5.360          \\
 & BLENDS & \textbf{2.509} & \textbf{0.757} & \textbf{3.746} & \textbf{4.408} \\
\hline
\multirow{4}{*}{Trajectory 2}
 & EKF    & 1.486          & 2.289          & 2.056          & 3.159          \\
 & TFS    & 1.350          & 2.344          & 1.736          & 3.090          \\
 & RTSS   & 1.332          & 2.283          & 1.796          & 3.078          \\
 & BLENDS & \textbf{0.499} & \textbf{1.803} & \textbf{1.716} & \textbf{2.351} \\
\hline
\end{tabular}
\end{table}

\section{Conclusions}\label{sec:conclusion}
\noindent
This paper presented BLENDS, a Bayesian learning-enhanced deep smoothing framework for INS/GNSS navigation in GNSS-denied environments. By integrating a transformer-based learning module within a classical smoothing framework, BLENDS learns to adapt the fusion of forward and backward estimates while preserving the statistical foundations of Bayesian estimation. Experimental results on a real-world quadrotor dataset demonstrated that the proposed approach consistently outperforms both the TFS and the RTSS, achieving substantial reductions in both position error and estimation uncertainty.
\\ \noindent
Beyond mitigating navigation drift during GNSS outages, BLENDS improves positioning accuracy beyond that achievable using conventional GNSS measurements alone. This improvement stems from the ability of the proposed framework to learn and compensate for the systematic bias that exists between standard GNSS positioning and the RTK-corrected reference solution, a limitation that cannot be addressed by classical smoothing methods. These results suggest that learning-enhanced smoothing offers a promising direction for achieving high-accuracy navigation using low-cost sensors without requiring expensive RTK infrastructure during operation.
\\ \noindent
Future work will investigate the application of BLENDS to longer GNSS outages, additional robotic platforms, and alternative sensor fusion architectures.
\bibliographystyle{IEEEtran}
\bibliography{bio.bib}

\end{document}